\documentclass[letterpaper, 10 pt, conference]{cls/ieeeconf}
\IEEEoverridecommandlockouts
\let\labelindent\relax

\usepackage[normalem]{ulem}	                        
\usepackage[table,usenames,dvipsnames]{xcolor}      
\usepackage{enumitem}                               
\usepackage{extarrows}                              
\usepackage[noadjust]{cite}                         

\usepackage{amsmath,amssymb,amsfonts,amsthm,dsfont} 
\usepackage{algorithm,algorithmicx,listings}        
\usepackage[noend]{algpseudocode}			        

\usepackage{graphicx,tabularx,subcaption}
\usepackage[export]{adjustbox}

\usepackage[font={small}]{caption}
\usepackage{multirow}

\usepackage[titletoc,title]{appendix}

\usepackage[breaklinks=true, colorlinks, bookmarks=true, citecolor=Black, urlcolor=Violet,linkcolor=Black]{hyperref}

\def\argmin{\mathop{\arg\min}\limits}

\DeclareMathOperator{\tr}{\mathbf{tr}}

\newcommand{\scaleMathLine}[2][1]{\resizebox{#1\linewidth}{!}{$\displaystyle{#2}$}}

\newcommand{\crl}[1]{\left\{#1\right\}}

\theoremstyle{definition}

\newtheorem*{assumption*}{Assumption}
\newtheorem*{problem*}{Problem}

\theoremstyle{remark}

\newtheorem*{solution*}{Solution}

\def\thetitle{Fully Convolutional Geometric Features for Category-level Object Alignment}
\def\theauthor{Qiaojun Feng, Nikolay Atanasov}
\def\thekeywords{}

\hypersetup{
  pdfauthor={\theauthor},%
  pdftitle={\thetitle},%
  pdfsubject={\thetitle},%
  pdfkeywords={\thekeywords}
}

\IEEEoverridecommandlockouts

\title{\LARGE \bf \thetitle}

\author{Qiaojun Feng \and Nikolay Atanasov
\thanks{We gratefully acknowledge support from ARL DCIST CRA W911NF-17-2-0181 and ONR SAI 00014-18-1-2828.}%
\thanks{The authors are with Department of Electrical and Computer Engineering, University of California, San Diego, La Jolla, CA 92093, USA {\tt\small \{qjfeng,natanasov\}@ucsd.edu}}
}

\begin{document}
\maketitle

\begin{abstract}
This paper focuses on pose registration of different object instances from the same category. This is required in online object mapping because object instances detected at test time usually differ from the training instances. Our approach transforms instances of the same category to a normalized canonical coordinate frame and uses metric learning to train fully convolutional geometric features. The resulting model is able to generate pairs of matching points between the instances, allowing category-level registration. Evaluation on both synthetic and real-world data shows that our method provides robust features, leading to accurate alignment of instances with different shapes.
\end{abstract}

\section{Introduction}
\label{sec:introduction}

\begin{figure}[t]
    \centering
    \includegraphics[width=\linewidth,trim=0mm 0mm 0mm 0mm, clip]{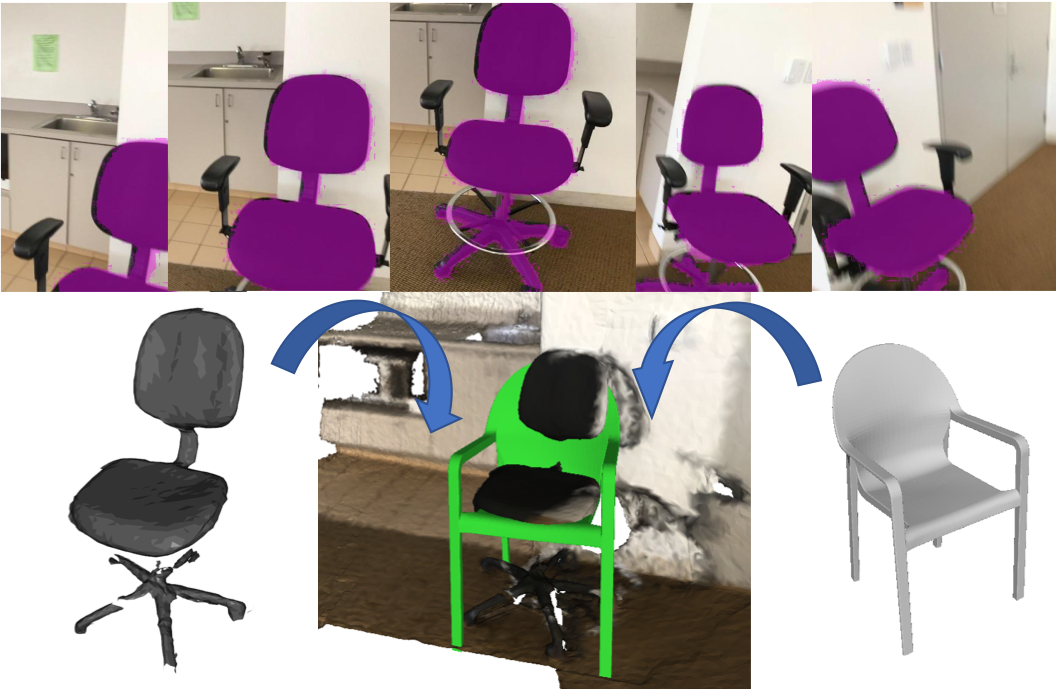}
    \caption{A sequence of RGB-D images (top) may be used for object-level mapping. This paper focuses on the important subproblem of category-level registration. A partial object point cloud (bottom, left) may be generated from object detection, segmentation, and tracking (top, purple). A not-identical CAD model (bottom, right) needs to be aligned to the observed point cloud (bottom, middle) in order to generate an object-level map.}
    \label{fig:intro}
\end{figure}

Meaningful and detailed environment reconstruction is an enabling capability for autonomous robot operation in various tasks, including safe navigation, object manipulation, or human-robot interaction. As embodied agents have limited storage and computation capabilities, developing compressed, yet, expressive environment models is a key problem. Online object-based simultaneous localization and mapping (SLAM) methods~\cite{slam++,semslam} are promising in generating efficient and semantically meaningful maps composed of sparse object landmarks. These methods work by detecting, segmenting and tracking object instances online and using the semantic information to estimate the object poses and shapes. Since the object instances observed online are always different from the stored models, semantic SLAM methods need to perform cross-instance alignment. Another challenge is that only partial and potentially low-resolution observations may be available due to the limited field-of-view or resolution of the onboard sensors. See Fig.~\ref{fig:intro} for an example.

Recently, learning-based 3D feature extraction algorithms have shown promising performance in point-cloud registration task. Convolutional neural network architectures can learn powerful descriptors from pre-aligned matching patches. However, generating matching patches for different instances in the same category may not be easy. While same-category objects share similar overall structure, individual parts may be quite different (e.g., armchair vs an office chair). Existing works rely on human annotations of sparse semantically meaningful object parts (e.g., chair legs, back support, seat) to obtain such matchings across instances~\cite{Pavlakos_Keypoints_ICRA2017}.

The main \textbf{contribution} of this paper is a learning-based method for generating dense matching pairs across different instances from the same category. We leverage the idea of normalized canonical coordinates (NCC)~\cite{Wang_NOCS_CVPR2019} to align the different instances during training and automatically generate positive and negative examples of matching pairs. We use metric learning to train a sparse convolutional neural network to predict cross-instance geometric features~\cite{Choy_FCGF_ICCV2019} suitable for registration. Our approach enables improved object pose estimation versus the state-of-the-art on both synthetic and real-world data.

\section{Related Work}
\label{sec:related_work}

\noindent\textbf{3D point cloud features:}
Traditional methods of hand-crafted 3D feature include Spin Image~\cite{Johnson_spin_PAMI1999}, FPFH~\cite{Rusu_FPFH_ICRA2009}, SHOT~\cite{Salti_SHOT_CVIU2014}. They mainly rely on simple local geometric information like normals and histogram. Recently learning-based methods have gained more attention. 
3DMatch~\cite{Zeng_3DMatch_CVPR2017} applies 3D ConvNet on fixed-size 3D patch represented in truncated distance function. 
3DFeat-Net~\cite{Yew_3DFeat_ECCV2018} collects local points in a radius-fixed ball, uses a detector module to estimate the local orientation and generates the descriptor after alignment. 
3DSmoothNet~\cite{Gojcic_Perfect_CVPR2019} proposes the idea of smoothed density value voxelization as the preprocessed input to reduce the sparsity of the input patch. 
FCGF~\cite{Choy_FCGF_ICCV2019} leverages the sparse convolution~\cite{Choy_Minkowski_CVPR2019} to build fully-convolutional network in 3D space and use metric learning to learn the feature. The point cloud coordinates and associated features are used for registration to estimate the rigid transformation.

\noindent\textbf{Pose estimation with known model:} 
Assume the object shape is given, we can estimate its 6DoF pose from RGB or RGB-D images. 
PVNet~\cite{Peng_PVNet_CVPR2019} estimates sparse object keypoints on the RGB image via voting mechanism. 
DPOD~\cite{Zakharov_DPOD_ICCV2019} predicts a dense object classification and 2D-3D correspondence mask. 
There are also algorithms doing a straight-forward pose regression without explicit associations such as \cite{Deng_PoseRBPF_RSS2019}. 
\cite{Gupta_Align_CVPR2015} trains a CNN to predict the object pose using pixel surface normals, followed by model selection and alignment using ICP. 
The Scan2CAD~\cite{Avetisyan_Scan2CAD_CVPR2019} dataset contains 9DoF alignment annotation of CAD models from ShapeNet~\cite{Chang_ShapeNet_2015} w.r.t. the indoor scene reconstruction of ScanNet~\cite{Dai_ScanNet_CVPR2017}. They voxelize the RGBD scan in the form of signed distance field (SDF). A 3D CNN network is trained to predict sparse keypoints correspondence with a fixed-size patch input.

\noindent\textbf{Category-level object pose estimation:} 
Objects in the same category usually share similar structure with relatively consistent distribution of semantic keypoints~\cite{Xiang_pascal3d+_WACV2014}, such as the wheels of the car and the chair legs. Category-level semantic keypoints~\cite{Pavlakos_Keypoints_ICRA2017} are predicted on RGB images and a deformable shape model is fitted to recover the pose. StarMap~\cite{Zhou_StarMap_ECCV2018} extends to predict category-agnostic keypoints by predicting the keypoint's normalized canonical coordinate from RGB image observation. NOCS~\cite{Wang_NOCS_CVPR2019} also uses the idea of normalized canonical coordinate and generate a dense annotation covering the whole object surface instead of sparse semantic keypoints. Our work is most similar to NOCS. The main difference is that we work on point cloud data and the point feature is predicted instead of the normalized coordinate.
\section{Problem Formulation}
\label{sec:problem_formulation}

Consider an RGBD camera moving in an unknown environment. Suppose that a convolutional neural network, such as Mask R-CNN~\cite{He_Mask_PAMI2020}, is used to detect and segment objects in each RGB image. Suppose that an object tracking algorithm, such as SiamMask~\cite{Wang_Track_CVPR2019}, tracks the detections over time. Suppose also that the camera pose is tracked using a SLAM algorithm, such as ORB-SLAM2~\cite{MurArtal_ORB_TRO2017}. Given the camera trajectory and the segmented RGBD pixels associated over time, we can construct a point cloud $\mathbf{X}\in\mathbb{R}^{N\times3}$ of the object in the world frame by accumulating the partial views and projecting them using the estimated camera poses.

Assume we have a small database of object models for the detected category $\mathcal{Y}=\{\mathbf{Y}_1,\dots,\mathbf{Y}_k\}$, where $\mathbf{Y}_i\in\mathbb{R}^{M_i\times3}$ is a point cloud. Given a query point cloud $\mathbf{X}$ observed online, we want to find a similar model $\mathbf{Y} \in \mathcal{Y}$ and estimate the pose $\mathbf{T} := [\mathbf{R}\;\mathbf{p};\;\mathbf{0}^{\top}\;1]\in SE(3)$ that aligns it with $\mathbf{X}$. Define a distance function between an incomplete point cloud $\mathbf{X}$ and complete point cloud $\mathbf{Y}$\ as $d(\mathbf{X},\mathbf{Y}): \mathbb{R}^{N\times3}\times\mathbb{R}^{M\times3} \to \mathbb{R}^{+}$. Our objective is to find an existing model $\mathbf{Y}^*$ and its pose $\mathbf{T}^*$ such that it fits an incomplete observation $\mathbf{X}$ well:
\begin{equation}
\label{eq:problem}
    \mathbf{Y}^*,\mathbf{T}^* = \argmin_{\mathbf{Y}\in \mathcal{Y},\mathbf{T}\in SE(3)} d(\mathbf{T}(\mathbf{X}),\mathbf{Y}),
\end{equation}
where the transformation is defined as:
\begin{equation}
\label{eq:transformation}
\scaleMathLine[0.9]{
    \mathbf{X} = \left[\mathbf{x}_{1} \; \cdots \; \mathbf{x}_{N} \right]^{\top}\;
    \mathbf{T}(\mathbf{X}) = \left[ \mathbf{R}\mathbf{x}_{1}+\mathbf{p} \; \cdots \; \mathbf{R}\mathbf{x}_{N}+\mathbf{p} \right]^{\top}}.
\end{equation}
\section{Background}

\subsection{Sparse Convolution}


Convolutional neural networks (CNNs) can be applied to 3D data directly by extending all 2D modules by one dimension. However, this not only increases the computation needed for training but also ignores the sparse structure of 3D data such as point clouds. Sparse convolution~\cite{Choy_Minkowski_CVPR2019} is an approach for efficient feature extraction on structured 3D data. Suppose $\mathbf{c}\in \mathbb{Z}^n$ is a coordinate in an $n$-dimensional quantized space and its associated feature vector is $\mathbf{x}_\mathbf{c}$. Define $\mathcal{V}^n(K)$ as the set of unit coordinate offsets around $\mathbf{0}$ in $n$-D space with size $K$ in each dimension. For example, $\mathcal{V}^1(3) = \{-1,0,1\}, \mathcal{V}^2(3) = \{\mathcal{V}^1(3) \times \mathcal{V}^1(3)\}$. Standard dense convolution combines features from all the region nearby:
\begin{equation}
    \mathbf{x}^{\text{out}}_{\mathbf{c}} =  \sum_{\mathbf{i}\in\mathcal{V}^n(K)} \mathbf{W}_\mathbf{i} \mathbf{x}^{\text{in}}_{\mathbf{c}+\mathbf{i}}.
\end{equation}
For sparse convolution, the selected offset $\mathbf{i}$ is based on the non-empty locations of the input. Suppose we can define the non-empty coordinate space $\mathcal{C}^{\text{in}}$ for $\mathbf{c}^{\text{in}}$ and $\mathcal{C}^{\text{out}}$ for $\mathbf{c}^{\text{out}}$. Then the sparse convolution is defined as
\begin{equation}
    \mathbf{x}^{\text{out}}_{\mathbf{c}} =  \sum_{\mathbf{i}\in \mathcal{U} (\mathbf{c},\mathcal{C}^{\text{in}})} \mathbf{W}_\mathbf{i} \mathbf{x}^{\text{in}}_{\mathbf{c}+\mathbf{i}} \; \text{for } \mathbf{c} \in \mathcal{C}^{\text{out}}
\end{equation}
, where $\mathcal{U} (\mathbf{c},\mathcal{C}^{\text{in}}) = \{\mathbf{i}|\mathbf{c}+\mathbf{i} \in \mathcal{C}^{\text{in}}\}$ only consider the existing adjacent feature value.

A fully convolutional network (FCN)~\cite{Long_FCN_CVPR2015} uses convolution and deconvolution to build the network structure without maxpooling or upsampling layers. FCN is suitable for dense prediction and its kernels have larger reception fields. The idea of fully convolutional layers can be combined with sparse convolution by using only convolutional layers.

\begin{figure}[t]
  \centering
  \includegraphics[width=\linewidth,trim=0mm 0mm 0mm 0mm, clip]{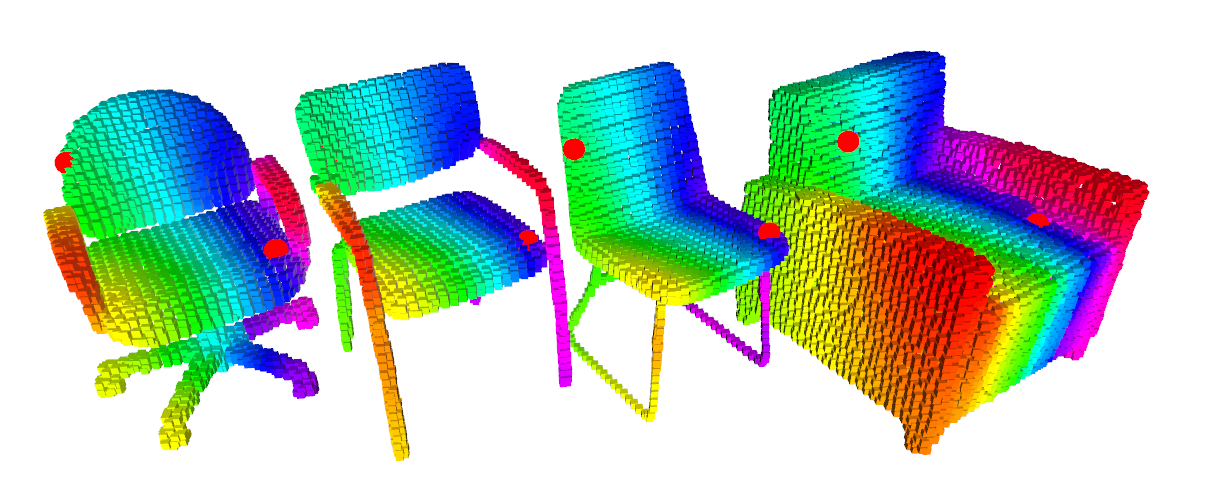}
  \caption{Visualization of different chairs in Normalized Canonical Coordinates (NCCs). Colors indicate the coordinates. The red dots annotate the same coordinate on different objects. Note that the red dot on the second chair is occluded by the back. Although not all points can be associated well in the NCCs, many positive matching pairs can be discovered.}
  \label{fig:NCC}
\end{figure}

\subsection{Normalized Canonical Coordinate}
The task of matching different objects is challenging. When scans are obtained from different object instance, the definition of good alignment becomes vague since a rigid transformation cannot eliminate the intrinsic shape difference. If the problem is restricted to aligning objects within the same category, we can leverage the conventions of object canonical pose. For example, a chair always has a seat and often a back and a canonical chair pose can be defined accordingly. The Normalized Object Coordinate Space (NOCS) is proposed in~\cite{Wang_NOCS_CVPR2019} and here we call this concept as Normalized Canonical Coordinate (NCC). The normalized object bounding box has a unit-length diagonal and is centered at the zero. Fig.~\ref{fig:NCC} visualizes some examples of the chairs in the category NCC. The NCC of a object can be recovered given its pose and scale. NCC can bridge different objects with different poses and shapes as a intermediate pose-invariant shape representation.

\section{Technical Approach}
\label{sec:technical_approach}

Given a source point cloud $\mathbf{X}$ and target point cloud $\mathbf{Y}$, we determine the transformation $\mathbf{T}$ that aligns $\mathbf{X}$ to $\mathbf{Y}$ in two steps. First, we extract features from both point clouds and establish point-to-point correspondences based on the features. Next, given the correspondences, we solve the alignment problem using a robust registration algorithm such as RANSAC~\cite{Fischler_RANSAC} or TEASER~\cite{teaser}.

\subsection{Feature Extraction Model}
\label{sec:feature_extraction}
Given a point cloud $\mathbf{X}$, we define a neural network model $f(\mathbf{X}; \boldsymbol{\theta})$ with parameters $\boldsymbol{\theta}$ to generate a set of point cloud features:
\begin{equation}
    \mathbf{F}_{\mathbf{X}} := f(\mathbf{X}; \boldsymbol{\theta}) = \crl{\mathbf{f}_1,\ldots,\mathbf{f}_N}
    \label{eq:feature}
\end{equation}
where $\mathbf{f}_i \in \mathbb{R}^k$ is the feature associated with point $\mathbf{x}_i$. We build a feature extractor using the ResUNet network proposed in FCGF~\cite{Choy_FCGF_ICCV2019}, which introduces residual blocks~\cite{He_ResNet_CVPR2016} 
into the UNet architecture~\cite{Ronneberger_UNet_MICCAI2015}
. One normalization layer is added before the output to generate unit-norm features. The network is trained using metric learning with training pairs generated as described in Sec.~\ref{sec:pair_matching}. We set the feature dimension to $k=32$.

\subsection{Pairs Matching}
\label{sec:pair_matching}

The main idea of metric learning is to learn a distance metric between objects in order to establish similarity or dissimilarity, which aligns with our goal. The goal of 3D point cloud feature learning is to find a feature space where matching pairs should be close to each other while non-matching pairs should be far apart. 

For the metric learning method, the generation of positive and negative pairs is at least as important as the loss design. 3DMatch~\cite{Zeng_3DMatch_CVPR2017} generates matching pairs between RGB-D scans from the camera poses recovered from different 3D reconstruction algorithm. 
Once two scans are aligned, positive pairs can be generated by local neighbor search, while the negative pairs are the remaining ones with larger distance. Define the positive matching pair set between two aligned point clouds $\mathbf{X},\mathbf{Y}$ as: 
\begin{equation}
    p(\mathbf{X},\mathbf{Y}) = \{(i,j) | \|\mathbf{x}_i-\mathbf{y}_j\|<\tau, i \leq N, j \leq M, i,j \in \mathbb{N}\},
    \label{eq:pairs}    
\end{equation}
where $\tau$ is the local neighbor search region. During training we usually only keep a subset of the positive matching pairs. Negative pairs are sampled from the complement set of the positive pair set.

\begin{figure}[t]
  \centering
  \includegraphics[width=\linewidth,trim=0mm 0mm 0mm 0mm, clip]{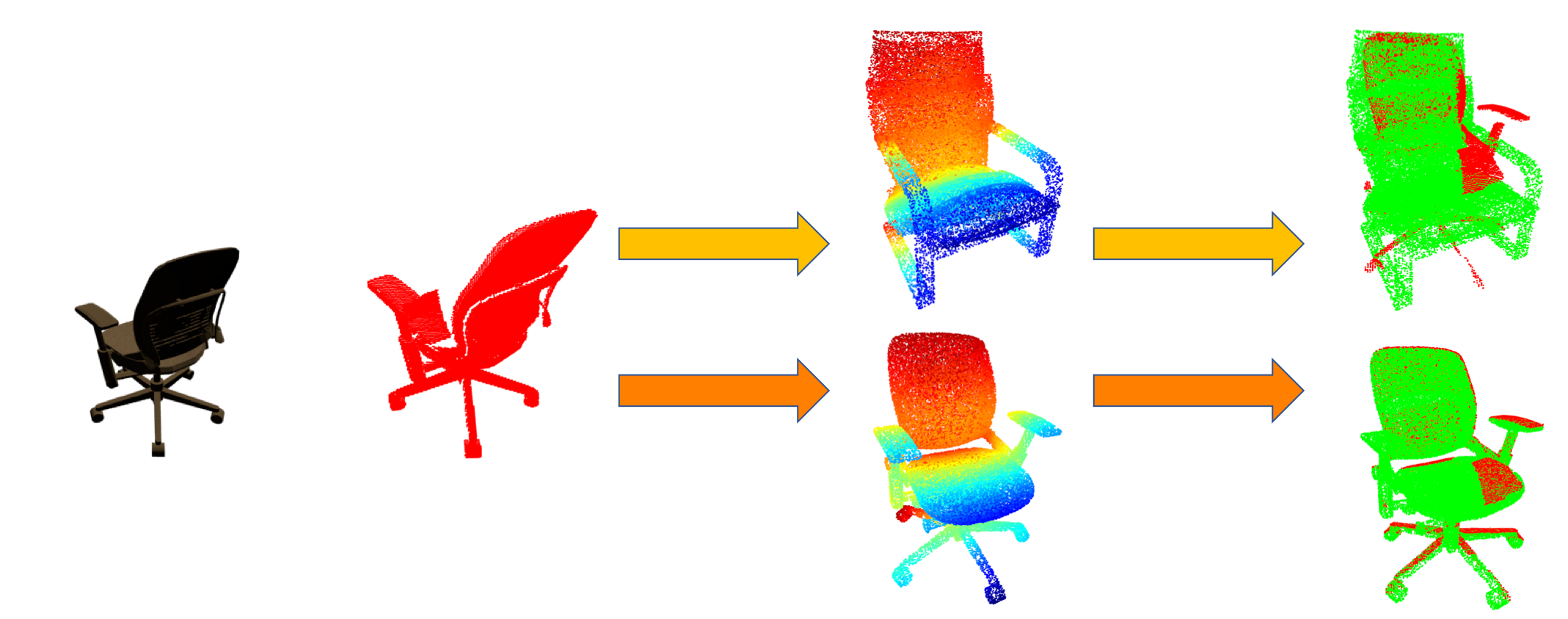}
  \caption{The observed point cloud (left) can be aligned with different CAD models (middle) in normalized canonical coordinate (right) to generate matching pairs. In the right column the observed is in red and the model is in green.}
  \label{fig:NCC_demo}
\end{figure}

We use the idea of NCC to bridge the connection between different object instances from the same category. 
We can convert an object point cloud $\mathbf{X}$ to the canonical frame with its pose $\mathbf{T}_\mathbf{X} \in SE(3)$ annotation and normalize it with the scale $s_\mathbf{X}\in\mathbb{R}$
\begin{equation}
    \mathbf{X}_{\text{NCC}} = \text{NCC}(\mathbf{X},s_\mathbf{X},\mathbf{T}_\mathbf{X}) = s_\mathbf{X}^{-1}\cdot\mathbf{T}_\mathbf{X}^{-1}(\mathbf{X})
    \label{eq:NCC}
\end{equation}
where $\mathbf{T}_\mathbf{X}^{-1}(\cdot)$ follows the same definition in eq.~(\ref{eq:transformation}). For two different object instances point clouds $\mathbf{X},\mathbf{Y}$, we convert them both into the NCC and find the positive matching pair set
\begin{equation}
    p'(\mathbf{X},\mathbf{Y}) = p(\mathbf{X}_{\text{NCC}},\mathbf{Y}_{\text{NCC}})
\end{equation}
as shown in Fig.~\ref{fig:NCC_demo}.

Our main intuition is that though we cannot get most of the points perfectly aligned, there are still enough roughly-aligned parts. Especially if we focus on the model pairs that are more similar than a random pick, we can generate quite a lot reasonable positive matching pair by extending the local neighbor search radius $\tau$ in eq.~(\ref{eq:pairs}). More details on matching pair generation and training on our customized dataset are introduced in Section~\ref{sec:experiments}.

\subsection{Metric Learning Loss}
\label{sec:metric_learning}

Unlike other learning-based 3D feature extractors which usually have some hand-crafted pre-processing step to concatenate some local feature, we uses fully-convolutional sparse convolution to automatically extract features. We define contrastive loss function~\cite{Hadsell_invariant_CVPR2006} used in metric learning for training. Assume $\mathbf{x},\mathbf{y}$ are two points from different but aligned point clouds $\mathbf{X},\mathbf{Y}$ and $i,j$ are their associated indices respectively. The associated features generate through feature extractor model $f(; \boldsymbol{\theta})$ in eq.~(\ref{eq:feature}) are $\mathbf{f}\mathbf{x},\mathbf{f}\mathbf{y}$.  $m_{\mathbf{x}\mathbf{y}}$ indicates the matching information. 
\begin{equation}
    m_{\mathbf{x}\mathbf{y}} = \left\{\begin{array}{cl}
        1, & \text{if } (i,j)\in p(\mathbf{X},\mathbf{Y})\\
        0, & \text{otherwise}
    \end{array}\right.
\end{equation}
See eq.~(\ref{eq:pairs}) for the definition of positive matching pairs $p(\mathbf{X},\mathbf{Y})$. Define $d(\cdot,\cdot)$ as a distance function between features. The contrastive loss is defined as
\begin{equation}
    L_{\text{con}}(\mathbf{f}\mathbf{x},\mathbf{f}\mathbf{y}) = m_{\mathbf{x}\mathbf{y}}(d(\mathbf{f}\mathbf{x},\mathbf{f}\mathbf{y})-p_{\text{pos}})^2 + \bar{m}_{\mathbf{x}\mathbf{y}}(d(\mathbf{f}\mathbf{x},\mathbf{f}\mathbf{y})-p_{\text{neg}})^2
\end{equation}
where $\bar{m}_{\mathbf{x}\mathbf{y}} = 1-m_{\mathbf{x}\mathbf{y}}$, $p_{\text{pos}},p_{\text{neg}}$ are distance threshold for positive and negative pair. These thresholds should be design such that the positive pairs move closer and the negative pairs get separated when the contrastive loss is decreasing. We set $p_{\text{pos}}=0.1,p_{\text{neg}}=1.4$ for our normalized feature vectors. And the 2-norm is used for distance function between features.

\subsection{Registration with Correspondences}
After the training, we have a neural network model as the feature extractor function $f(\cdot; \boldsymbol{\theta})$. 
We fixing the $\boldsymbol{\theta}$, use the feature extraction model for inference and extract point-wise features $\mathbf{F}_{\mathbf{X}},\mathbf{F}_{\mathbf{Y}}$ for the point clouds $\mathbf{X},\mathbf{Y}$ as in eq.~(\ref{eq:feature}). We can perform the nearest-neighbor search for each point in $\mathbf{X}$ to generate correspondences, which is similar to eq.~(\ref{eq:pairs}) but calculate the Euclidean distance between feature vectors instead of point coordinates. 
\begin{equation}
\begin{aligned}
    p_f(\mathbf{F}_{\mathbf{X}},\mathbf{F}_{\mathbf{Y}}) = \{(i,j_i) | j_i=\arg\min_j d(\mathbf{f}{\mathbf{x}_i},\mathbf{f}{\mathbf{y}_j}), \\
    i \leq N, j_i \leq M, i,j_i \in \mathbb{N}\}
\end{aligned}
    \label{eq:feature_pairs}    
\end{equation}

We then solve the pose estimation, or the point cloud registration problem with correspondences using robust methods like RANSAC~\cite{Fischler_RANSAC}. In each iteration, RANSAC randomly samples $k$ pairs of matching points $\{(\mathbf{x}_i,\mathbf{y}_i)\}_{i=1}^k$ and solve the minimization problem
\begin{equation}
    \mathbf{T}^* = \arg\min_{T\in SE(3)} \sum_{i=1}^k \|T(\mathbf{x}_i) - \mathbf{y}_i\|^2
\end{equation}
using Kabsch algorithm~\cite{Kabsch_1976}. Then the $\mathbf{T}^*$ is evaluate on the whole set to see how many inliers it contains as the consensus set. The estimation with largest consensus set is maintained. We use the RANSAC implementation in Open3D~\cite{Zhou_Open3D_2018}.

\section{Experiments}
\label{sec:experiments}


\begin{figure*}[t]
  \centering
  \includegraphics[width=0.48\linewidth,trim=0mm 0mm 0mm 0mm, clip]{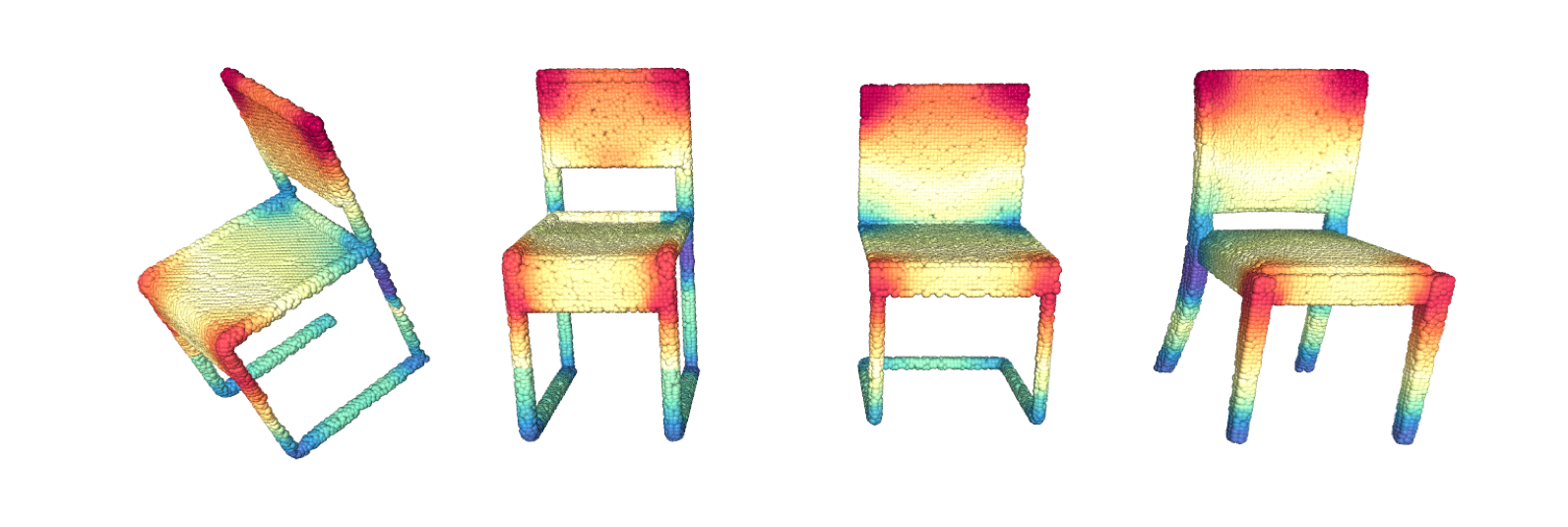}
  \includegraphics[width=0.48\linewidth,trim=0mm 0mm 0mm 0mm, clip]{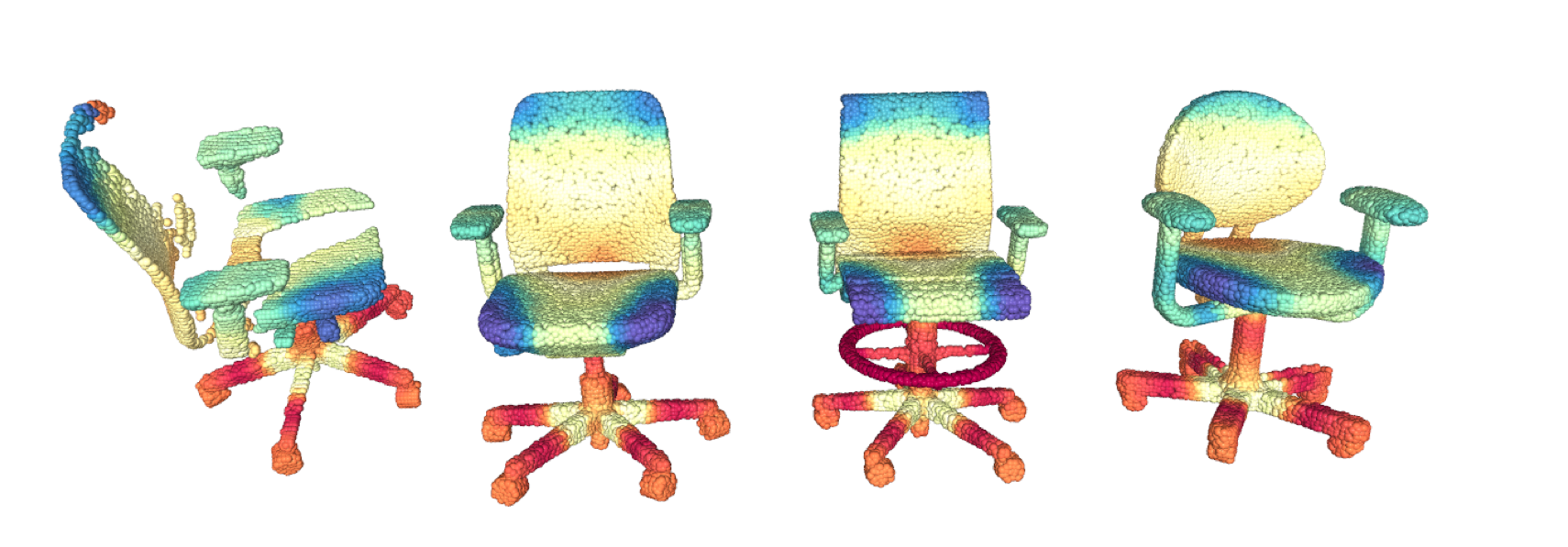}
  \caption{Examples of point feature embedded by t-SNE. Column 1: pointcloud from single depth scan. Column 2: identical CAD model. Column 3\&4: neighbor models.}
  \label{fig:feature_color}
\end{figure*}

\begin{figure*}[t]
  \centering
  \includegraphics[width=\linewidth,trim=0mm 0mm 0mm 0mm, clip]{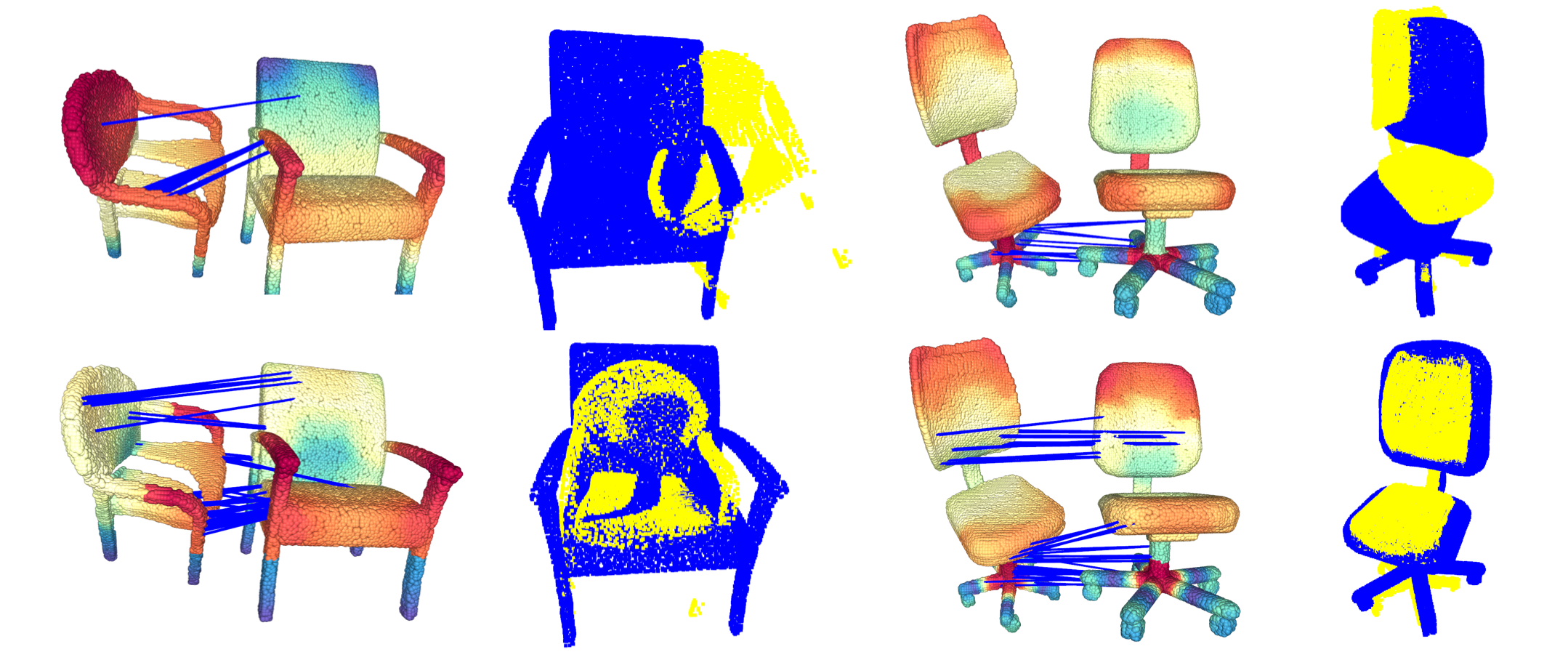}
  \caption{Comparing the models trained with (bottom row) and without (top row) cross-instance matching data on synthetic data. Column 1 and 3 show the point feature embedded by t-SNE. The depth scan is on the left and the CAD model is on the right. The blue lines indicate the accurate match among the total 1000 sampled matches. Column 2 and 4 show the final registration results based on 1000 matching pairs. The CAD model is in blue and the depth scan is in yellow. The negative matching pairs are not shown, which cause the pose estimation failure.}
  \label{fig:vis_syn_compare}
\end{figure*}

\subsection{Dataset}
Our dataset is built from Scan2CAD~\cite{Avetisyan_Scan2CAD_CVPR2019}. We focus on the category of chair and we select the scenes from subcategory of Lobby and Conference Room in ScanNet~\cite{Dai_ScanNet_CVPR2017}. Based on the appearance annotations of Scan2CAD we collect 137 chair models for training and 42 chair models for testing, both from ShapeNet~\cite{Chang_ShapeNet_2015}. We build a synthetic dataset from these ShapeNet models in canonical pose. In the synthetic dataset, for each CAD model we render depth images from 10 fixed viewpoints 
and convert them into point clouds. 

To prepare for the non-identical model matching, we decide to annotate the model neighboring pairs instead of matching each pair of them. This step acts as providing an oracle for choosing the $\mathbf{Y}^*$ in eq.~(\ref{eq:problem}). We choose the 3-nearest-neighbors for each model's point cloud based on the Earth Mover's distance (EMD)~\cite{Fan_PointSet_CVPR2017}. For two point sets $\mathbf{X},\mathbf{Y}$ with same size, the Earth Mover's distance is defined as
\begin{equation}
    d_{EMD}(\mathbf{X}, \mathbf{Y}) = \min_{\Phi:\mathbf{X} \to \mathbf{Y}} \sum_{\mathbf{x} \in \mathbf{X}}\|\mathbf{x} - \Phi(\mathbf{x})\|_2
\end{equation}
where $\Phi: \mathbf{X} \to \mathbf{Y}$ is a one-to-one correspondence. We sample 2048 points on each CAD model to calculate the EMD. For both training and testing set the neighbors are annotated. Notice for the CAD models in the testing set their neighbors are selected only in the training set. 

For the real-world data, we look into the scene reconstruction meshes in ScanNet. From the 117 training scenes we segment out 974 chair meshes and convert them into point clouds. Also for the 45 testing scenes 331 chairs are selected. For each of these chair the Scan2CAD provides its associated ShapeNet CAD model and its pose. We use this to evaluate the performance of real-world point cloud pose estimation. Besides, some RGB-D images with object segmentations are collected for quantitative evaluation in Sec.~\ref{sec:experiments_real}.

\subsection{Evaluation Metric}

We borrow the idea of matched fragments from~\cite{Deng_PPFNet_CVPR2018} and evaluate the accuracy of matching pairs. Assume $\mathbf{X}$ and $\mathbf{Y}$ are two associated point clouds with the matching between the same index. The groundtruth transformation from $\mathcal{X}$ to $\mathcal{Y}$ is $\mathbf{T}^*$. The matching accuracy evaluates how many matching pairs can be aligned within some threshold $\tau_1$ after applying the groundtruth transformation.
\begin{equation}
    \text{MatchAcc}(\mathbf{X},\mathbf{Y}) = \frac{1}{n}\sum_{i=1}^n\mathds{1}\left(\|\mathbf{T}^*(\mathbf{x}_i)-\mathbf{y}_i\|_2 < \tau_1\right)
\end{equation}
Here we set $\tau_1=0.05\;m$. In~\cite{Deng_PPFNet_CVPR2018} the authors set the positive inlier ratio to be $0.05$, indicating that as long as there are $5\%$ of the matching pairs we can get a good registration result using some robust algorithms to remove the outliers. 

For cross-instance matching we cannot have a general matching distance due to the shape variations. Complementarily, we can also estimate the relative pose and measure the relative error. The pose can be represented by the rotation part and translation part. Define the groundtruth pose as $(\mathbf{R}^{*},\mathbf{p}^{*})$ and the estimated pose as $(\mathbf{\hat{R}},\mathbf{\hat{p}})$, where $\mathbf{R}^{*},\mathbf{\hat{R}} \in SO(3), \mathbf{p}^{*},\mathbf{\hat{p}} \in \mathbb{R}^3$. We can decouple them to define the relative rotation error (RRE)
\begin{equation}
    \text{RRE}(\mathbf{\hat{R}},\mathbf{{R}^{*}}) = \arccos\{[\tr{(\mathbf{\hat{R}}^T \mathbf{R}^{*})}-1]/2\}
\end{equation}
and the relative translation error (RTE)
\begin{equation}
    \text{RTE}(\mathbf{\hat{p}},\mathbf{{p}^{*}}) = \|\mathbf{\hat{p}}-\mathbf{p^{*}}\|_2
\end{equation}



\begin{figure}[t]
  \centering
  \includegraphics[width=0.49\linewidth,trim=0mm 0mm 15mm 0mm, clip]{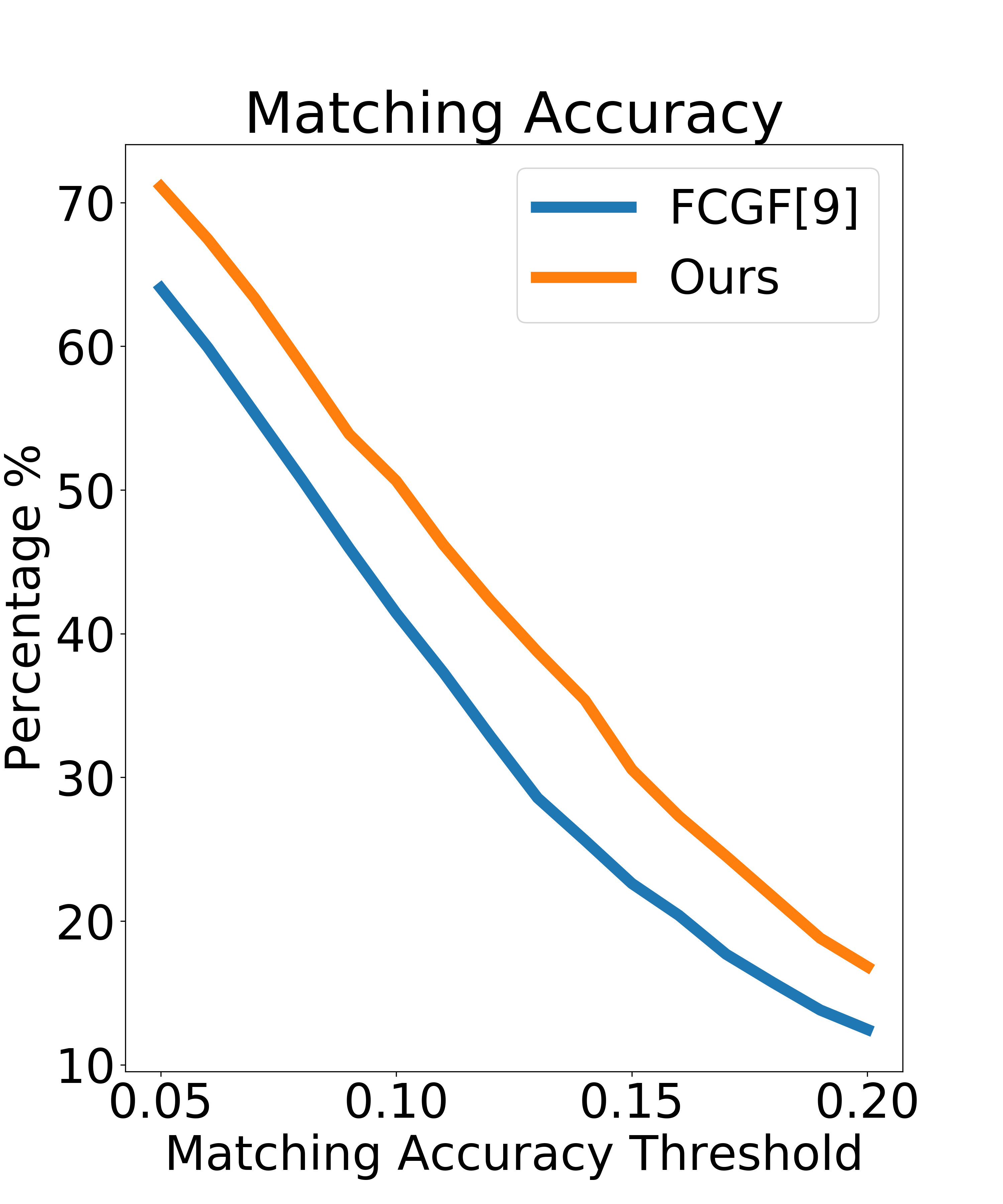}
  \includegraphics[width=0.49\linewidth,trim=0mm 0mm 15mm 0mm, clip]{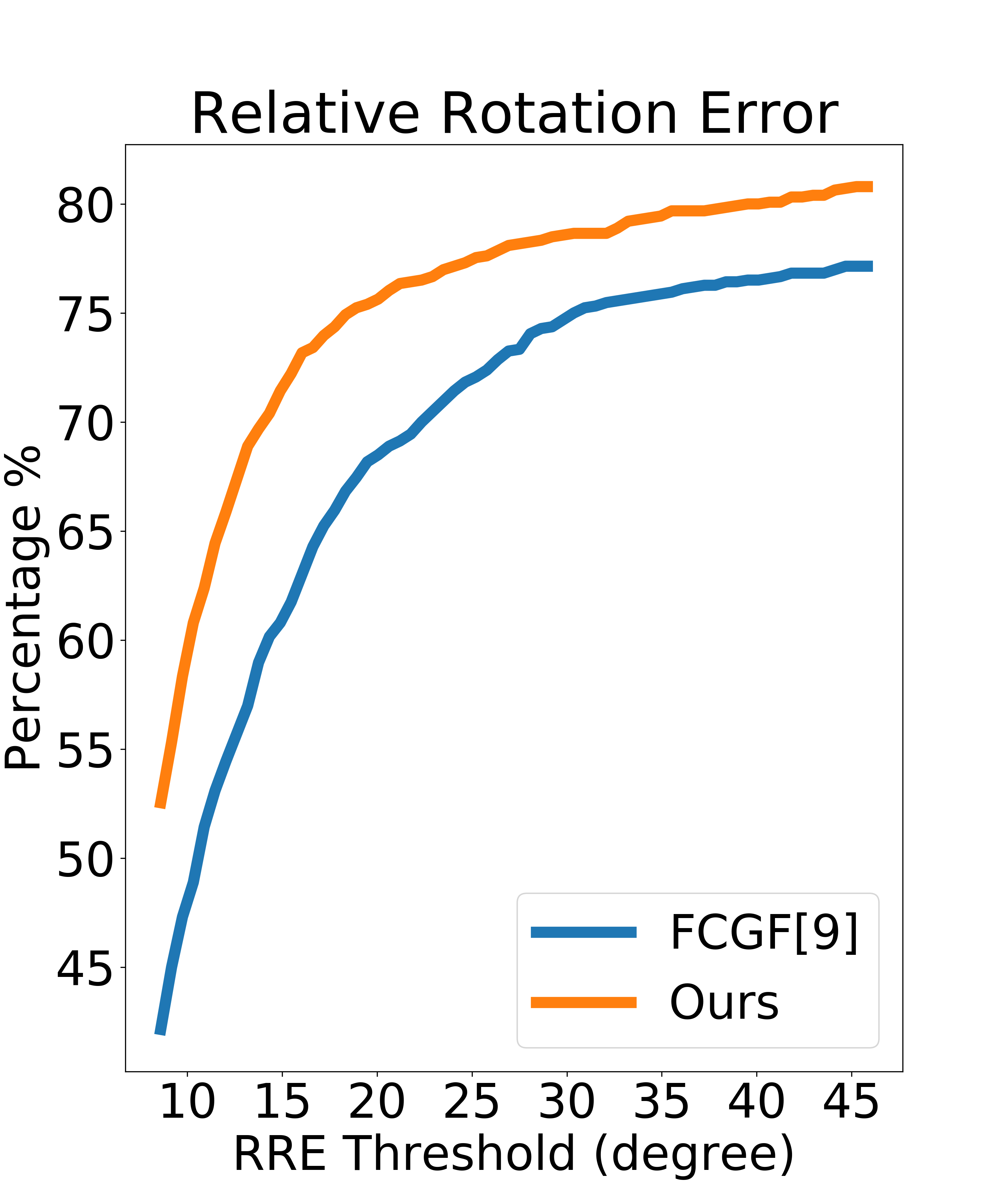}
  \caption{Results on ShapeNet synthetic data. Left: matching accuracy. Right: relative rotation error. FCGF~\cite{Choy_FCGF_ICCV2019} is the baseline model w/o cross-instance matching.}
  \label{fig:exp_syn}
\end{figure}

\begin{figure}[t]
  \centering
  \includegraphics[width=0.49\linewidth,trim=0mm 0mm 17mm 0mm, clip]{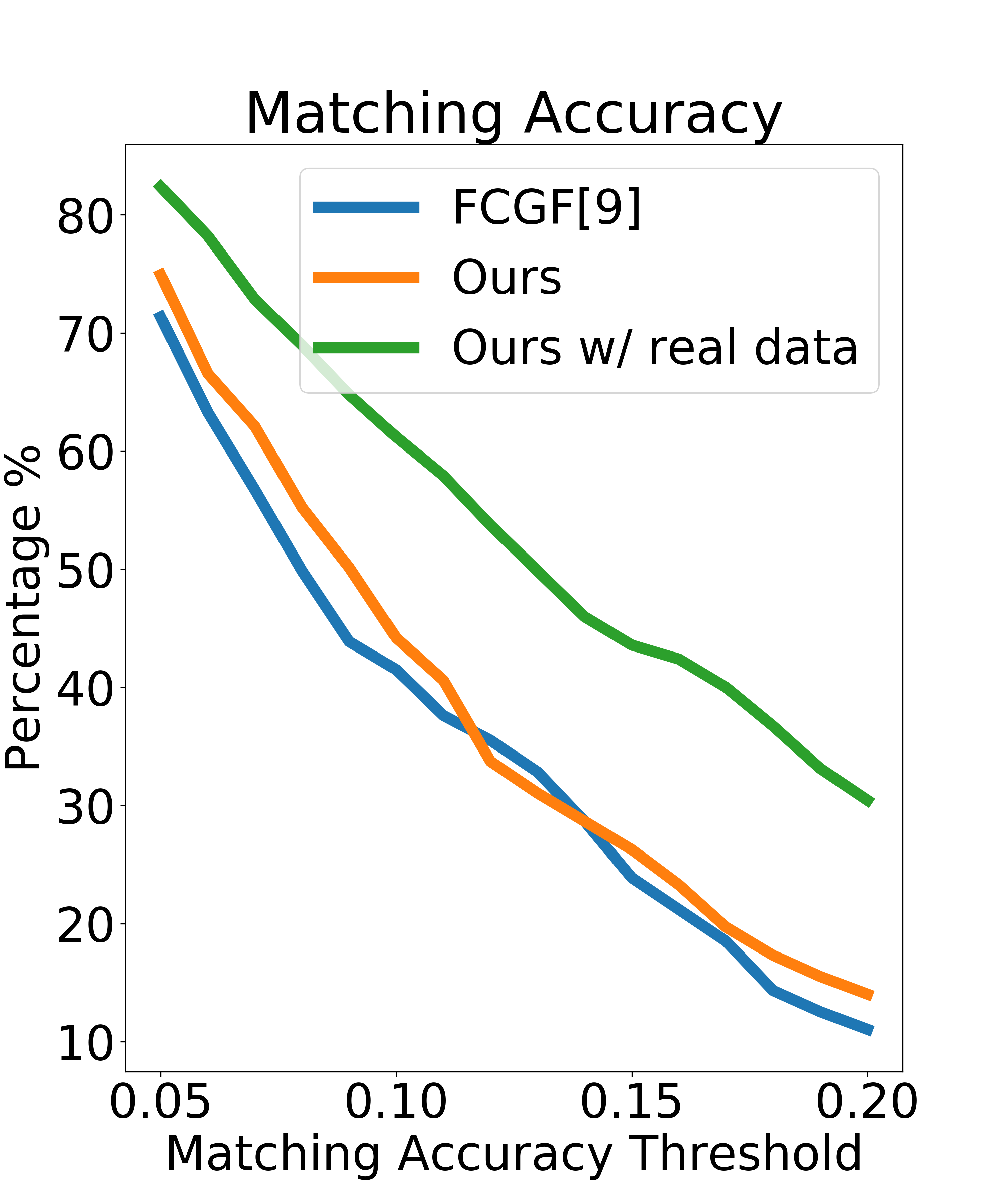}
  \includegraphics[width=0.49\linewidth,trim=0mm 0mm 17mm 0mm, clip]{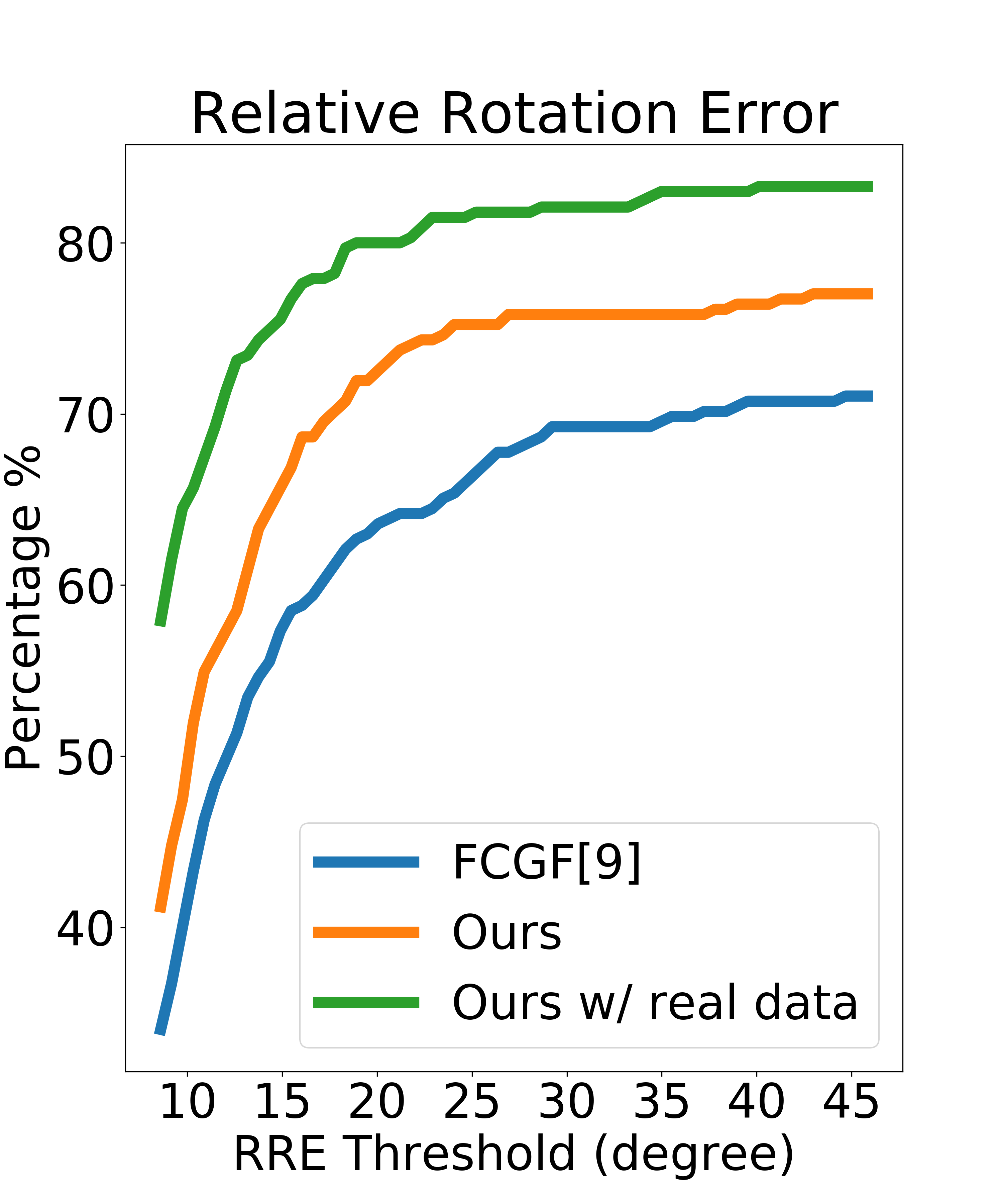}
  \caption{Results on ScanNet real-world data. Left: matching accuracy. Right: relative rotation error. FCGF~\cite{Choy_FCGF_ICCV2019} is the baseline model w/o cross-instance matching. Our w/ real data is the model trained with real-world point clouds.}
  \label{fig:exp_real}
\end{figure}

\subsection{Synthetic Dataset: ShapeNet}
From the synthetic dataset, each CAD model in the training set and its 10 point clouds observations at different poses are used for model training. We use the curriculum learning~\cite{Bengio_Curriculum_ICML2009} to verify our cross-instance matching data augmentation idea so that the training procedure is more stable. We first train on the same-instance pairs. Each CAD model can pair with its own incomplete point clouds to generate 10 same-instance pairs. The baseline model is trained on the pairs for 100 epochs. 

Afterwards we introduce cross-instance matching pairs for the training. Based on the EMD neighbors we compute, we generate pairs between each CAD model and their EMD-neighboring CAD models. Also we pair the incomplete depth image point cloud with its EMD-neighboring CAD models. We keep a few same-instance pairs as used in the baseline to make the training more stable. The model with cross-instance matching is trained for another 100 epochs. For a fair comparison, we also keep training the baseline model for another 100 epochs solely on same-instance pairs and name it as the model without cross-instance matching.

By pairing with EMD neighbor we can discover more meaningful and stable matching pairs. And we may derive implicit connections with more than the neighbors since the neighbor CAD also has its own neighbor. Notice that here the neighbor CADs are all included in the original training set. We are not introducing new data but generate new matching pairs in the original database. This can be viewed as a data augmentation strategy for this specific feature learning task. 
In Fig.~\ref{fig:feature_color} we visualize the point cloud features between the partial point cloud and the CAD models by embedding the high dimensional feature vector using t-SNE~\cite{Maaten_tSNE_JMLR2008} and coloring the point cloud.

During testing, the incomplete point clouds associated with the CAD models in the test set are used as observation. We assume that we don't have access to the test models that generate these observation. Instead, we have the pre-selected EMD neighbors of each test CAD model from the training set. 
For each observed point cloud, three models from the training set is provided as candidates and we use the trained model for the task of cross-instance matching and pose estimation. The comparison result is shown in Fig.~\ref{fig:exp_syn} and Table.~\ref{tab:RRE_RTE}. We visualize some comparing cases on pose estimation in Fig.~\ref{fig:vis_syn_compare}. 
We can see that with cross-instance matching data included for the training, the model can generate more consistent feature vectors across different instance, estimate more inlier matching pairs and also perform better on the pose estimation task.

\begin{figure}[t]
  \centering
  \includegraphics[width=\linewidth,trim=0mm 0mm 0mm 0mm, clip]{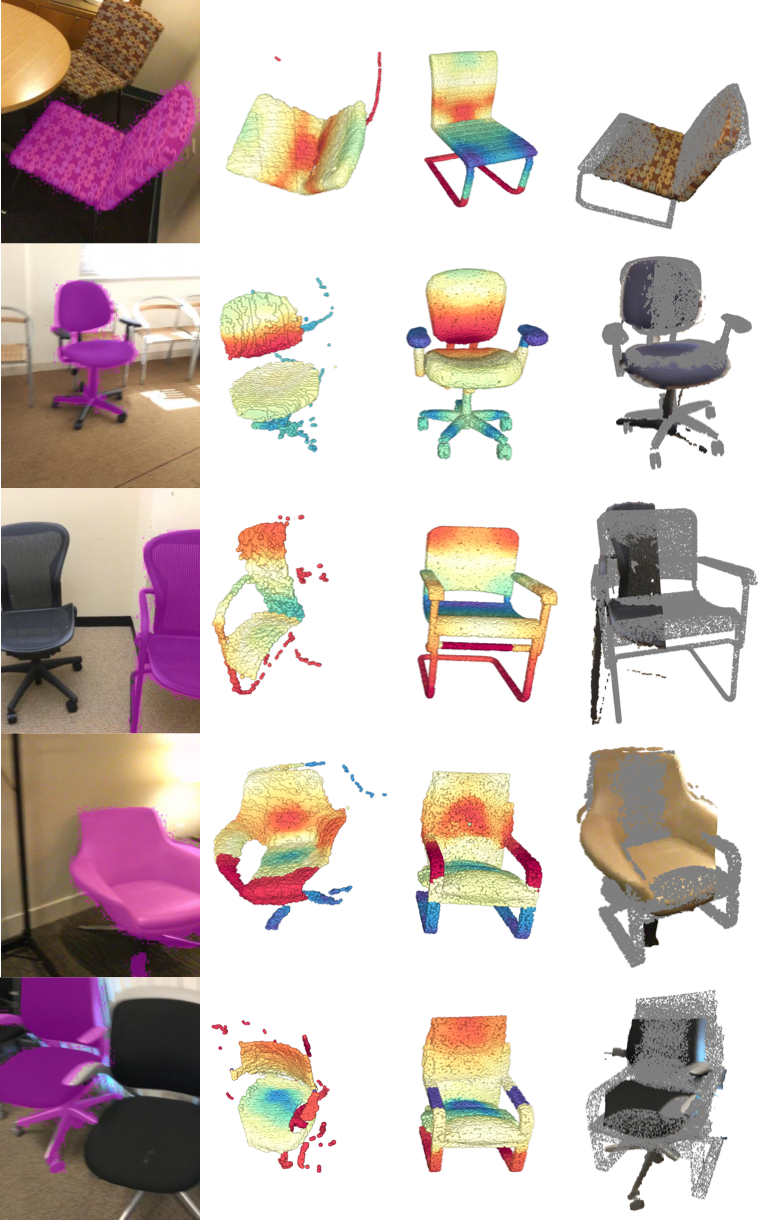}
  \caption{Examples on matching the point cloud segmented from a single RGB-D image. Column 1: RGB images and object segmentatin masks in purple. Column 2 \& 3: color-coded point cloud features of the observed objects and the CAD models. Column 4: Alignment results with the CAD models painted in grey. }
  \label{fig:real_pos}
\end{figure}

\begin{figure}[t]
  \centering
  \includegraphics[width=\linewidth,trim=0mm 0mm 0mm 0mm, clip]{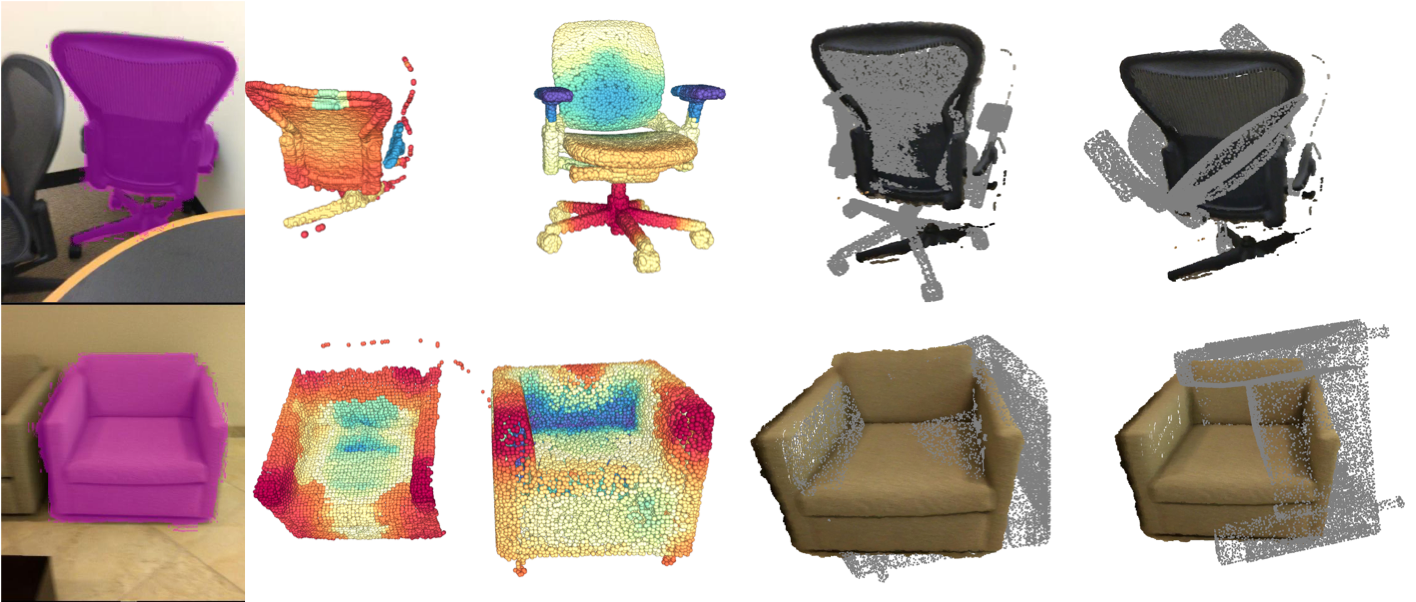}
  \caption{Failure cases on RGB-D point cloud registration. Column 1: RGB images and object segmentatin masks in purple. Column 2 \& 3: color-coded point cloud features of the observed objects and the CAD models. Column 4 \& 5: Different alignment results after running RANSAC, with the CAD models painted in grey.}
  \label{fig:real_neg}
\end{figure}

\begin{table}[t]
  \caption{The percentage of the testing data within the RRE/RTE threshold. The RRE values are aligned with Fig.~\ref{fig:exp_syn}, \ref{fig:exp_real}. Ours+ is the model trained on the real-world data.}
  \label{tab:RRE_RTE}
  \centering
  \begin{tabular}{cc|ccc|cc}
        \hline
        \multirow{2}{*}{Dataset} &  \multirow{2}{*}{Method} & \multicolumn{3}{c|}{RRE (degree)} & \multicolumn{2}{c}{RTE (cm)} \\
		 & & 10 & 20 & 30 & 5 & 10 \\
		\hline
		\multirow{2}{*}{Synthetic} & FCGF\cite{Choy_FCGF_ICCV2019} & 48.02 & 68.49 & 74.84 & 47.38 & 72.78 \\
		& Ours & \textbf{59.21} & \textbf{75.56} & \textbf{78.57} & \textbf{55.32} & \textbf{78.49}\\
		\hline
		\multirow{3}{*}{Real-world} & FCGF\cite{Choy_FCGF_ICCV2019} & 41.49 & 63.58 & 69.25 & 45.97 & 68.06 \\
		& Ours & 50.15 & 72.54 & 75.82 & 54.33 & 74.33 \\
		& Ours+ & \textbf{65.07} & \textbf{80.00} & \textbf{82.09} & \textbf{69.55} & \textbf{83.88}\\
		\hline
  \end{tabular}
\end{table}

\subsection{Real-world Dataset: ScanNet}
\label{sec:experiments_real}
For the real-world dataset we are trying to match the CADs with the segmented point cloud from the scene reconstruction mesh. Here we only use the annotated CAD model from Scan2CAD but not the neighbor models because the potential sim-real domain gap. We do a transfer learning, training on the basis of the model learned on the synthetic dataset which has already learned from cross-instance data. Fig.~\ref{fig:exp_real} is showing the performance of different models and quantitative results are listed in Table.~\ref{tab:RRE_RTE}. With cross-instance matching training the model can have a better performance on the real-world data even there is domain transfer gap and the model has never seen real-world data. Also with transfer learning on the real-world data the model can have higher matching accuracy and the lower rotation and translation error.

We visualize some qualitative result in Fig.~\ref{fig:real_pos}. Here we extract the observed object point cloud from a segmented RGB-D image. The first and second rows show examples with relatively complete observations. The third example is incomplete due to image truncation. The fourth and the fifth observed objects are sharing the same CAD model for pose estimation. Most of them are showing reasonable performance. There are some failure cases shown in Fig.~\ref{fig:real_neg}. When the inlier matching pairs are limited, the RANSAC algorithm provides unstable estimation. We observe that observation from the back of the chair is challenging for the model because it is very similar to a plane without specific geometric structure. Also sometimes the pose estimation of cube-shape sofa chair is noisy. One possible reason is that there are too much outlier matching pairs on the plane surfaces. One potential solution to this is to train a classifier to justify useful points and their features for pose estimation instead of using all of them or randomly sampling a subset.

\section{Conclusion}
\label{sec:conclusion}

In this work, we perform the category-specific cross instance point cloud matching and pose estimation. We propose to convert different instances from the same category into the normalized canonical coordinate to make a good alignment, which is helpful for positive matching pairs generation in metric learning. We train a fully-convolutional sparse convolution model for point cloud feature extraction. During testing, we use RANSAC to estimate the pose with associations from the features. Our future work will focus on the prediction and optimization in the shape-deformation space. It is interesting to investigate more on the problem of CAD model retrieval given partial observations. Given a relative good pose and shape initialization, we want to refine the object pose and shape jointly. 


{\small
\bibliographystyle{cls/IEEEtran}
\bibliography{bib/ref.bib}
}
\end{document}